\documentclass{article}

\usepackage[preprint]{neurips_2025}

\usepackage[utf8]{inputenc} 
\usepackage[T1]{fontenc}    
\usepackage{hyperref}       
\usepackage{url}            
\usepackage{booktabs}       
\usepackage{amsfonts}       
\usepackage{nicefrac}       
\usepackage{microtype}      
\usepackage{xcolor}         
\usepackage{amsthm}
\usepackage{amssymb}  
\usepackage{pifont}
\usepackage{amsmath}
\usepackage{algorithm}
\usepackage{algpseudocode} 
\usepackage{graphicx}
\newtheorem{theorem}{Theorem}[section] 
\newtheorem{definition}{Definition}
\newtheorem{example}{Example}
\newtheorem{proposition}[theorem]{Proposition} 

\usepackage{tikz}

\title{Autoregressive Meta-Actions for\\ Unified Controllable Trajectory Generation}

\author{
  Jianbo Zhao\thanks{Equal contribution.} \quad
  Taiyu Ban\footnotemark[1] \\
  University of Science and Technology of China \\
  96 Jinzhai Rd, Hefei 230026, China. \\
  \texttt{\{zjb123,banty\}@mail.ustc.edu.cn}
  \And
  Xiyang Wang\footnotemark[1] \\
  Mach Drive \\
  Suzhou Rd 3, Beijing 100080, China\\
  \texttt{\{xiyang.wang\}@mach-drive.com}\\
  \And
  Qibin Zhou \quad Hangning Zhou\thanks{Project leader.} \quad Zhihao Liu \quad Mu Yang \\
  Mach Drive \\
  Suzhou Rd 3, Beijing 100080, China\\
  \texttt{\{qibin.zhou,hangning.zhou,zhihao.liu,mu.yang\}@mach-drive.com}\\
  \And
  Lei Liu\thanks{Corresponding author.} \quad Bin Li \\
  University of Science and Technology of China \\
  96 Jinzhai Rd, Hefei 230026, China.\\
  \texttt{\{liulei13,binli\}@ustc.edu.cn}\\
}

\begin{document}

\maketitle

\begin{abstract}

Controllable trajectory generation guided by high-level semantic decisions, termed meta-actions, is crucial for autonomous driving systems. A significant limitation of existing frameworks is their reliance on invariant meta-actions assigned over fixed future time intervals, causing temporal misalignment with the actual behavior trajectories. This misalignment leads to irrelevant associations between the prescribed meta-actions and the resulting trajectories, disrupting task coherence and limiting model performance. To address this challenge, we introduce Autoregressive Meta-Actions, an approach integrated into autoregressive trajectory generation frameworks that provides a unified and precise definition for meta-action-conditioned trajectory prediction.
Specifically, We decompose traditional long-interval meta-actions into frame-level meta-actions, enabling a sequential interplay between autoregressive meta-action prediction and meta-action-conditioned trajectory generation. This decomposition ensures strict alignment between each trajectory segment and its corresponding meta-action, achieving a consistent and unified task formulation across the entire trajectory span and significantly reducing complexity.
Moreover, we propose a staged pre-training process to decouple the learning of basic motion dynamics from the integration of high-level decision control, which offers flexibility, stability, and modularity. Experimental results validate our framework's effectiveness, demonstrating improved trajectory adaptivity and responsiveness to dynamic decision-making scenarios. We provide the  video document available at \url{https://arma-traj.github.io/}.

\end{abstract}

\section{Introduction}

Trajectory generation \citep{seff2023motionlm, zhang2024trafficbots, multipath} guided by high-level semantic decisions \citep{huang2024versatile, hu2023planning}, termed meta-actions \citep{sima2024drivelm, hwang2024emma}, is essential for achieving controllability and interpretability in autonomous driving systems. Meta-actions, such as lane keeping, lane changes, and turns, represent strategic driving behaviors, providing an abstract semantic framework that simplifies the interpretation and control of vehicle trajectories. To effectively learn these abstractions, it is crucial to ensure a precise temporal alignment between each meta-action and its corresponding trajectory segment. 

Conventionally, controllable trajectory generation frameworks assign a single meta-action label \citep{hu2023planning, jiang2023vad, hwang2024emma} over fixed, relatively long temporal intervals, and then learn trajectory patterns corresponding to these intervals. However, such long-interval labeling inevitably introduces irrelevant associations between meta-actions and their trajectory segments. Specifically, trajectory intervals that \textit{span transition points between different meta-actions} are forced into association with a \textit{single meta-action}, causing semantic misalignment. Consequently, this misalignment disrupts the unified semantic objective of consistently learning decision-aligned trajectories throughout the entire timeline. Furthermore, the resulting noisy and semantically inconsistent trajectory space substantially increases the difficulty of modeling decision-following behavior.

Recently, autoregressive models \citep{wu2024smart, zhang2025closed, lin2025revisit} have achieved notable success in trajectory generation by treating frame-level trajectories as discrete tokens and decomposing long-horizon trajectory generation into sequential, step-by-step predictions of the next frame. Building upon this foundation, some studies incorporate meta-actions into autoregressive frameworks \citep{rowe2024ctrl, zhang2025carplanner} to guide and control agent motions. However, these methods still condition the generation process on fixed, long-interval meta-actions, thus retaining the fundamental limitation of semantic misalignment between meta-actions and trajectory segments, and failing to achieve a unified task formulation.

To address these limitations, we propose the Autoregressive Meta-Action framework, integrated within the autoregressive trajectory generation framework. Rather than relying on fixed, long-interval meta-actions, we decompose high-level decisions \citep{bousdekis2021review} into frame-level meta-actions, ensuring that each frame's trajectory strictly aligns with its corresponding meta-action. On this basis, we reformulate the task into a sequential interplay between an autoregressive meta-action prediction process, which enables forward generation of meta-actions, and a meta-action-conditioned trajectory generation process, which enables controllability.

Specifically, we first predict the next-frame meta-action based on historical trajectories and meta-action sequences. We localize each frame within the temporal progression of the ongoing decision, allowing the model to accurately detect transitions between meta-actions. This enables the model to flexibly and responsively adjust high-level behaviors in accordance with evolving environmental interactions. Subsequently, the predicted meta-action conditions the trajectory generation process, aligning trajectory patterns with their corresponding meta-actions.
In this task definition, frame-level trajectories and meta-actions remain consistently aligned throughout the entire prediction horizon, achieving a unified and semantically coherent formulation for controllable trajectory generation.

Practically, we propose a staged fine-tuning framework that enables plug-and-play integration of meta-actions into a backbone model trained for generalizable trajectory generation. Firstly, we train a foundational autoregressive model solely on trajectory data, without meta-action supervision, allowing it to learn general motion patterns and environment interactions. In the second stage, we freeze the parameters of this foundation model and introduce two additional modules: (1) a meta-action prediction module and (2) a meta-action injection module that conditions trajectory generation on high-level decisions. Only these new modules are fine-tuned, enabling stable and modular incorporation of specific decision-following capabilities. If new decision types need to be integrated, we can fine-tune the meta-action-related modules alone, without retraining the entire model.

Finally, we present a method for generating frame-level meta-action labels spanning the entire temporal extent of trajectories, enabling straightforward adaptation from conventional long-interval task definitions to our proposed autoregressive formulation. Additionally, we provide an open-source experimental dataset derived from the Waymo Motion dataset \citep{sun2020scalability}, augmented with comprehensive frame-level meta-action annotations.

Sections of related work, label generation of frame-level meta-actions, supplementary model details, and supplementary experiemental results are available in the appendix.



\section{Preliminaries}
This section introduces existing formulations of controllable trajectory generation. We begin with the conventional regression-based setting, followed by the autoregressive framework.

\paragraph{Controllable Trajectory Generation}
Let \( \mathbf{S}_{1:t} = \{ \mathbf{s}_1, \mathbf{s}_2, \ldots, \mathbf{s}_t \} \) denote the sequence of historical agent states (e.g., positions, velocities, headings) up to the current time \( t \), where \( \mathbf{s}_i \) is the state at time step \( i \). Let \( \mathbf{E}_{1:t} \) represent the observed environment context over time (e.g., surrounding agents, map elements, traffic signals). Let \( c \in \mathcal{C} \) be a high-level meta-action describing the intended behavior (e.g., lane change, turn, stop), drawn from a finite semantic action set \( \mathcal{C} \).

The goal is to learn a function \( f \) that maps the input history and decision to a future trajectory:
\begin{equation}
    \hat{\mathbf{S}}_{t+1:t+T} = f(\mathbf{S}_{1:t}, \mathbf{E}_{1:t}, c),
\end{equation}
where \( \hat{\mathbf{S}}_{t+1:t+T} = \{ \hat{\mathbf{s}}_{t+1}, \ldots, \hat{\mathbf{s}}_{t+T} \} \) is the predicted trajectory over a horizon of \( T \) steps. This is often cast probabilistically as:
\begin{equation}
    P(\hat{\mathbf{S}}_{t+1:t+T} \mid \mathbf{S}_{1:t}, \mathbf{E}_{1:t}, c).
    \label{eq:regression_condition_task}
\end{equation}

If the system supports predicting the future meta-action $c$ without a strict dependency on externally provided inputs, the task can be reformulated as a joint prediction of the trajectory and meta-action:
\begin{equation}
P(\hat{\mathbf{S}}_{t+1:t+T}, c \mid \mathbf{S}_{1:t}, \mathbf{E}_{1:t}) = P(\hat{\mathbf{S}}_{t+1:t+T} \mid \mathbf{S}_{1:t}, \mathbf{E}_{1:t}, c)\, P(c \mid \mathbf{S}_{1:t}, \mathbf{E}_{1:t}).
\label{eq:regressive_overall_task}
\end{equation}
We adopt this formulation with predictive meta-actions due to its practical flexibility.

\paragraph{Autoregressive Framework}
The autoregressive setting decomposes Equation~\eqref{eq:regressive_overall_task} by predicting each future state sequentially, while conditioning on a fixed meta-action:
\begin{equation}
    P(\hat{\mathbf{S}}_{t+1:t+T}, c \mid \mathbf{S}_{1:t}, \mathbf{E}_{1:t})
= \left( \prod_{\tau = t+1}^{t+T} P(\hat{\mathbf{s}}_\tau \mid \mathbf{S}_{1:\tau-1}, \mathbf{E}_{1:\tau-1}, c) \right) P(c \mid \mathbf{S}_{1:t}, \mathbf{E}_{1:t}).
\label{eq:autoregressive_condition_task}
\end{equation}
Here, the fixed decision \( c \) guides the prediction at every future step. This reflects the standard assumption in classical controllable trajectory generation that agent behavior remains consistent over the prediction interval. However, this simplification can lead to temporal misalignment between the fixed decision and the actual transition in behavior, as will be analyzed later.

\section{Unified Task Formulation with Autoregressive Meta-Actions}
This section first analyzes the lack of temporal unification in existing task formulations for controllable trajectory generation, and then introduces our unified formulation using autoregressive meta-actions.

\subsection{Non-Unification of Existing Task Formulations}
We begin by defining what it means for a temporal task to be unified.

\begin{definition}[Task Unification]
    Let \( f: \mathcal{X}_t \rightarrow \mathcal{Y}_t \) be the function to be learned in a temporal task, where \( \mathcal{X}_t \) and \( \mathcal{Y}_t \) denote the input and output spaces at time \( t \). We say that \( f \) is unified over a temporal sequence \( \{t_1, \dots, t_N\} \) if it satisfies:
    \begin{enumerate}
        \item \textbf{Structural Consistency:} \(\forall\, t_i, t_j \in \{t_1, \dots, t_N\},\, \mathcal{X}_{t_i} = \mathcal{X}_{t_j} \) and \( \mathcal{Y}_{t_i} = \mathcal{Y}_{t_j} \); that is, the input and output structures are consistent across all time steps.
        \item \textbf{Semantic Consistency:} \(\forall\, t \in \{t_1, \dots, t_N\},\, f(\mathcal{X}_t) \) has the same functional interpretation.
    \end{enumerate}
\end{definition}

Consider the conventional formulation in Equation~\eqref{eq:regression_condition_task}\footnote{Equation~\eqref{eq:regression_condition_task} is the decision-conditioned subcomponent of the broader formulation in Equation~\eqref{eq:regressive_overall_task}.}. This formulation fails to ensure semantic consistency in controllable trajectory generation, which is intended to map the input history and a given meta-action to a future trajectory that faithfully follows that meta-action. We illustrate this failure in the following example (see also Figure~\ref{fig:ma_example}):

\begin{figure}[!h]
    \centering
    \includegraphics[width=1.\linewidth]{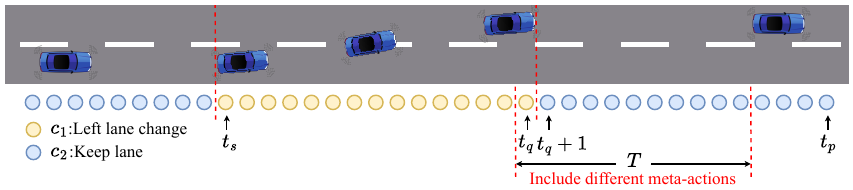}
    \caption{Example of different meta-actions in a sliding window with interval $T>1$.}
    \label{fig:ma_example}
\end{figure}

\begin{example}
    Let the time interval \((t_s, \ldots, t_q)\) be associated with meta-action \( c_1 \), and the subsequent interval \((t_{q+1}, \ldots, t_p)\) be associated with meta-action \( c_2 \). Suppose the prediction horizon is \( T > 1 \), and consider the task defined by Equation~\eqref{eq:regression_condition_task} at time \( t = t_{q-1} \). The predicted future trajectory \((t_q, t_{q+1}, \ldots, t_{q+T-1})\) overlaps both meta-action segments \( c_1 \) and \( c_2 \). In this case, the function \( f \) attempts to map the input and meta-action \( c_1 \) to a future trajectory that is partially governed by a different meta-action \( c_2 \), thereby violating semantic consistency.
    \label{example:not_unify}
\end{example}

This example reveals the semantic misalignment inherent in the conventional formulation\footnote{Note that the autoregressive version in Equation~\eqref{eq:autoregressive_condition_task} also suffers from the same unification issue.}.

\begin{proposition}
    The regression-based task formulation defined in Equation~\eqref{eq:regression_condition_task} is not unified with respect to mapping input histories and meta-actions to future trajectories.
\end{proposition}

\paragraph{Necessity of Frame-Level Meta-Actions}
Notably, Example \ref{example:not_unify} provides the insight that, as long as the future time interval $T>1$ (longer than the unit interval), we can always find a predicted trajectory segment containing at least two different meta-actions.
This insight highlights the \textit{necessity of frame-level representation} of meta-actions for a unified task formulation, introduced subsequently.

\subsection{Task Formulation with Autoregressive Meta-Actions}
To address the lack of unification in existing task definitions, we represent meta-actions at unit time intervals and propose a unified autoregressive framework for controllable trajectory generation.

Formally, let \( \mathbf{S}_{1:t} = \{ \mathbf{s}_1, \mathbf{s}_2, \ldots, \mathbf{s}_t \} \) denote the agent's historical trajectory, and \( \mathbf{E}_{1:t} \) the corresponding environmental context. We define frame-level meta-actions as \( \mathbf{C}_{1:t} = \{ \mathbf{c}_1, \mathbf{c}_2, \ldots, \mathbf{c}_t \} \), where \( \mathbf{c}_\tau \) represents the active meta-action at time \( \tau \) that guides the generation of $s_{\tau+1}$.
The task is defined to jointly predict future trajectories and their associated meta-actions in an autoregressive fashion:
\begin{subequations}
\begin{align}
    &P(\hat{\mathbf{S}}_{t+1:t+T}, \hat{\mathbf{C}}_{t:t+T-1} \mid \mathbf{S}_{1:t}, \mathbf{E}_{1:t}, \mathbf{C}_{1:t-1}) \label{eq:task_defintion_a}\\
    &= \prod_{\tau=t}^{t+T-1} P(\hat{\mathbf{s}}_{\tau+1}, \hat{\mathbf{c}}_{\tau} \mid \mathbf{S}_{1:\tau}, \mathbf{E}_{1:\tau}, \mathbf{C}_{1:\tau-1}) \label{eq:task_defintion_b}\\
    &= \prod_{\tau=t}^{t+T-1}
    P(\hat{\mathbf{c}}_{\tau} \mid \mathbf{S}_{1:\tau}, \mathbf{E}_{1:\tau}, \mathbf{C}_{1:\tau-1}) \,
    P(\hat{\mathbf{s}}_{\tau+1} \mid \mathbf{S}_{1:\tau}, \mathbf{E}_{1:\tau}, \mathbf{C}_{1:\tau}). \label{eq:task_defintion_c}
\end{align}
\label{eq:task_defintion}
\end{subequations}

This formulation treats meta-actions and trajectories with the same structural granularity, enabling joint prediction over time. 
Equation~\eqref{eq:task_defintion_b} decomposes the joint distribution autoregressively over time. At each time step \( \tau \), we predict the active meta-action $\hat{\mathbf{c}}_{\tau}$ and the resulted next-frame state \( \hat{\mathbf{s}}_{\tau+1} \), then update the history for the next step.
Equation~\eqref{eq:task_defintion_c} further factorizes this prediction into two stages: first, predicting the next meta-action \( \hat{\mathbf{c}}_\tau \); then, generating the trajectory state \( \hat{\mathbf{s}}_{\tau+1} \) conditioned on the newly predicted meta-action. This factorization follows the causal relationship in driving behavior, where high-level decisions determine motion outcomes.
By externally fixing the meta-action $\hat{\mathbf{c}}_\tau$ at specific steps, one can directly imposes targeted intervention or user-defined guidance to the model.

\paragraph{Incorporating Historical Meta-Action Sequences}
Our formulation includes the history of meta-actions as part of the input, enabling the model to better localize the temporal phase of the current decision. In contrast, conventional formulations (Equations~\eqref{eq:regressive_overall_task} and~\eqref{eq:autoregressive_condition_task}) typically omit this information, relying on an implicit and idealized alignment between trajectory segments and high-level decisions. As illustrated in Example~\ref{example:not_unify}, such alignment is often inaccurate in practice.
By explicitly incorporating historical meta-action sequences\footnote{Since frame-level meta-actions are inherently aligned with trajectory segments, removing the condition on meta-action history does not break semantic consistency. But for model performance, it is necessary to include this information to reduce ambiguity and facilitate learning decision boundaries.}, we eliminate the need for the model to infer decision boundaries solely from trajectory data, thereby reducing modeling complexity.
This operation facilitates the learning of decision transitions via autoregressive modeling of meta-actions, enabling flexible and responsive switching of high-level behaviors in response to environmental interactions.

\paragraph{Unification of the Task Formulation}
We now analyze the semantic unification of the proposed formulation in Equation~\eqref{eq:task_defintion_c}. The autoregressive meta-action prediction component,

$$
P(\hat{\mathbf{c}}_{\tau} \mid \mathbf{S}_{1:\tau}, \mathbf{E}_{1:\tau}, \mathbf{C}_{1:\tau-1}),
$$

defines a consistent mapping from historical trajectories, environmental context, and prior meta-actions to the prediction of the next meta-action $\hat{\mathbf{c}}_{\tau}$.

In parallel, the meta-action-conditioned trajectory generation process,
$$
P(\hat{\mathbf{s}}_{\tau+1} \mid \mathbf{S}_{1:\tau}, \mathbf{E}_{1:\tau}, \mathbf{C}_{1:\tau}),
$$
consistently maps the input history and current meta-action $\hat{\mathbf{c}}_{\tau} \in \mathbf{C}_{1:\tau}$ to the frame-level trajectory state $\hat{\mathbf{s}}_{\tau}$ that corresponds to this currently active meta-action.

Importantly, the semantics of both sub-processes remain consistent at every time step $\tau$, thereby satisfying both structural and semantic consistency across the temporal sequence. This confirms that the formulation achieves task unification.

\begin{proposition}
The autoregressive task formulation defined by Equation~\eqref{eq:task_defintion_c} is unified with respect to meta-action prediction and meta-action-conditioned trajectory generation.
\end{proposition}

This result establishes that our autoregressive meta-action framework constitutes a unified and coherent formulation for controllable trajectory generation. In the next section, we detail the model architecture designed to implement this formulation.

\begin{figure}[!h]
    \centering
    \includegraphics[width=0.99\linewidth]{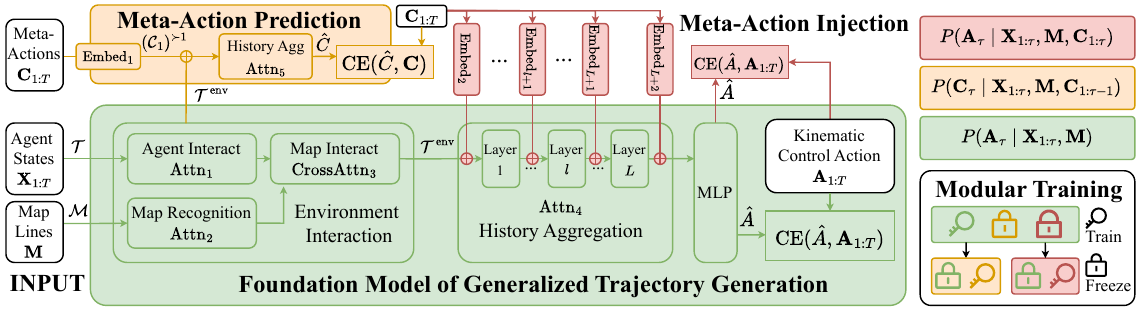}
    \caption{The overall diagram of the proposed model architecture.}
    \label{fig:framework}
\end{figure}

\section{Model Architecture}
We propose a modular architecture for autoregressive meta-action-based controllable trajectory generation (Equation~\eqref{eq:task_defintion_c}), as shown in Figure \ref{fig:framework}. The model consists of a foundation trajectory generation module that captures general motion and interaction patterns, a meta-action prediction module that autoregressively infers future decisions, and a meta-action injection module that integrates these decisions to guide trajectory generation. The foundation model is first trained independently, while the latter two modules are fine-tuned with its parameters frozen.

\paragraph{Symbol Notation}
Our model follows a query-centric design, treating all agents uniformly without explicitly separating ego and surrounding agents and thus requiring a re-clarification of inputs.
Let \( \mathbf{X}_t = \{ \mathbf{x}_t^{(i)} \}_{i=1}^{N} \) denote the set of agent states (which consist of the agent's position, velocity, heading and acceleration) at time \( t \), where \( \mathbf{x}_t^{(i)} \in \mathbb{R}^d \) is the state vector of agent \( i \). Similarly, let \( \mathbf{C}_t = \{ \mathbf{c}_t^{(i)} \in \{1, 2, \ldots, d_c\} \}_{i=1}^{N} \) denote the set of meta-actions associated with each agent at time \( t \), where \( d_c \) is the number of defined meta-action classes. The map context is represented by \( \mathbf{M} = \{ \mathbf{m}^{(k)} \}_{k=1}^K \), where each \( \mathbf{m}^{(k)} \in \mathbb{R}^{d_m} \) encodes a map polyline or lane segment.

For notation involving multi-dimensional slicing, we use \( \mathbf{T}_{i_1:j_1,\, i_2:j_2,\, \dots} \) to indicate selection along specific dimensions of a tensor \( \mathbf{T} \), where \( i_k:j_k \) defines the index range on the \( k \)-th axis.
This notation also applies to sets, which are treated as one-dimensional vectors.

\subsection{Foundation Trajectory Generation Model}
We build upon the kinematic-based trajectory representation introduced by \citet{zhao2024kigras} for the foundation model. In this representation, transitions between consecutive trajectory states are governed by underlying physical control actions, following the form  
\[
\mathbf{x}_{\tau+1}^{(i)} = \mathcal{K}(\mathbf{x}_{\tau}^{(i)}, \mathbf{a}_{\tau}^{(i)}),
\]
where \( \mathcal{K} \) is a kinematic model \citep{werling2010optimal}, and \( \mathbf{a}_{\tau}^{(i)} \) denotes the (kinematic) control action consisting of acceleration and yaw rate.
The control action sequence at each time step \( t \) is denoted as \( \mathbf{A}_t = \{ \mathbf{a}_t^{(i)} \in \{1, \dots, d_a\} \} \), where \( d_a \) is the number of discrete bins defined by the joint discretization of the control action space. These actions are derived from the observed state transitions \( \mathbf{X}_{t:t+1} \).

Autoregressive trajectory generation is thus formulated as predicting autoregressive control actions:
\begin{equation}
   \prod_{\tau=t}^{t+T-1} P(\mathbf{A}_{\tau} \mid \mathbf{X}_{1:\tau}, \mathbf{M}), \quad \text{where the agent states are updated as } \mathbf{x}_{\tau+1}^{(i)} = \mathcal{K}(\mathbf{x}_{\tau}^{(i)}, \mathbf{a}_{\tau}^{(i)}).
\end{equation}
To model interactions and temporal dependencies, we apply multi-head attention for both agent interaction and historical aggregation. Let \( \mathcal{T} \in \mathbb{R}^{N \times T \times D} \) denote the tokenized representation of agent states \( \mathbf{X}_{1:T} \), where \( N \) is the number of agents, \( T \) is the temporal window size, and \( D \) is the embedding dimension. Similarly, let \( \mathcal{M} \in \mathbb{R}^{K \times D} \) represent the tokenized map context, where each of the \( K \) polylines is encoded into a \( D \)-dimensional vector.

For environment interaction modeling, we first apply self-attention over the agent tokens \( \mathcal{T} \) along the agent dimension to capture agent-agent interactions. Next, we process the map tokens \( \mathcal{M} \) via self-attention to encode map structure, followed by cross-attention from agent tokens to map tokens to capture agent-map interactions. All attention operations are equipped with rotary position embeddings (RoPE) \citep{su2024roformer}, with implementation details provided in Appendix~\ref{sec:app_model_details}.

\begin{subequations}
    \begin{gather}
        \mathcal{T}^{\text{agent}} = \text{Attn}^{\text{RoPE}}_1\left(
        \mathbf{Q} = \mathcal{T}_{:,\tau},\ 
        \mathbf{K} = \mathcal{T}_{:,\tau},\ 
        \mathbf{V} = \mathcal{T}_{:,\tau}
        \right),
        \label{eq:cross_atten_agent_agent}\\
        \mathcal{M}^{\text{map}} = \text{Attn}^{\text{RoPE}}_2\left(
        \mathbf{Q} = \mathcal{M}_{:},\ 
        \mathbf{K} = \mathcal{M}_{:},\ 
        \mathbf{V} = \mathcal{M}_{:}
        \right),
        \label{eq:cross_atten_map_map}\\
        \mathcal{T}^{\text{env}} = \text{Attn}^{\text{RoPE}}_3\left(
        \mathbf{Q} = \mathcal{T}^{\text{agent}}_{:,\tau},\ 
        \mathbf{K} = \mathcal{M}^{\text{map}}_{:},\ 
        \mathbf{V} = \mathcal{M}^{\text{map}}_{:}
        \right).
        \label{eq:cross_atten_agent_map}
    \end{gather}
    \label{eq:cross_atten_agent}
\end{subequations}

To model temporal dependencies, we apply a causal self-attention mechanism along the time dimension for each agent, aggregating its historical states:
\begin{equation}
    \mathcal{T}^{\text{history}} = \text{Attn}^{\text{RoPE}}_4\left(
    \mathbf{Q} = \mathcal{T}_{i,\tau}^{\text{env}},\ 
    \mathbf{K} = \mathcal{T}_{i,1:\tau}^{\text{env}},\ 
    \mathbf{V} = \mathcal{T}_{i,1:\tau}^{\text{env}}
    \right).
    \label{eq:cross_atten_time}
\end{equation}

The history-aggregated tokens are passed through an MLP to decode the control action distribution:
\begin{equation}
    \hat{A} = \text{MLP}(\mathcal{T}^{\text{history}}_{1:N,:}), \quad 
    \text{where } \hat{A} \in \mathbb{R}^{N \times T \times d_a} \text{ denotes the predicted control action logits}.
    \label{eq:ffn}
\end{equation}

\paragraph{Training} 
The foundation model is trained using the cross-entropy (CE) loss between the predicted control distributions \( \hat{A} \) and the ground-truth discrete control actions \( \mathbf{A}_{1:T} \):
\begin{equation}
    \mathcal{L}_{\text{found}} = \text{CE}(\hat{A}, \mathbf{A}_{1:T}).
\end{equation}

\subsection{Meta-Action Prediction Module}

To incorporate decision history into meta-action prediction, we first embed the meta-action sequence \( \mathbf{C}_{1:T} \) and apply a one-step temporal shift. The shifted embeddings are then added to the agent tokens \( \mathcal{T}^{\text{env}} \), obtained from the environment interaction module in Equation~\eqref{eq:cross_atten_agent}:

\begin{gather}
    \mathcal{C}_1 = \text{Embed}_1(\mathbf{C}_{1:T}), \quad \mathcal{C}_1 \in \mathbb{R}^{N \times T \times D}, \\
    \mathcal{T}^{\text{c-env}} = \mathcal{T}^{\text{env}}_{:,1:T} + (\mathcal{C}_1)^{\succ 1}, \label{eq:meta-action-shift}
\end{gather}

where \( (\mathcal{C}_1)^{\succ 1} = [\mathbf{0},\, (\mathcal{C}_1)_{1:T-1}] \) denotes the meta-action embeddings shifted forward by one timestep, with a zero vector prepended along the temporal axis.

\paragraph{Causal Contitioning} This shift ensures that the agent token at time step \( t \) has access only to its past meta-actions \( \mathbf{C}_{1:t-1} \), thereby avoiding information leakage from the future. It preserves the autoregressive prediction setting, where each step is conditioned only on historical information.

We then apply causal self-attention along the temporal dimension to aggregate historical contexts:

\begin{equation}
    \mathcal{T}^{\text{c-history}} = \text{Attn}^{\text{RoPE}}_5\left(
    \mathbf{Q} = \mathcal{T}^{\text{c-env}}_{i,\tau},\ 
    \mathbf{K} = \mathcal{T}^{\text{c-env}}_{i,1:\tau},\ 
    \mathbf{V} = \mathcal{T}^{\text{c-env}}_{i,1:\tau}
    \right).
\end{equation}

The history-aggregated representation is then decoded into meta-action predictions:

\begin{equation}
   \hat{C} = \text{FFN}(\mathcal{T}^{\text{c-history}}), \quad
   \text{where } \hat{C} \in \mathbb{R}^{N \times T \times d_c} \text{ denotes the predicted meta-action logits}.
\end{equation}

\paragraph{Critical Timing of Meta-Action Injection}
A key design decision in Equation~\eqref{eq:meta-action-shift} is the precise timing at which meta-action embeddings are introduced. This choice is essential for preserving strict alignment with the task formulation.
If meta-action embeddings are injected too early (i.e., before the environment interaction step \( \mathcal{T}^{\text{env}} \)), they will leak across agents via the attention mechanism in Equation~\eqref{eq:cross_atten_agent}, violating the fact that each agent’s meta-actions are private and invisible to other agents during prediction.
Conversely, injecting meta-actions after history aggregation \( \mathcal{T}^{\text{history}} \) diminishes their effectiveness, since the relevant contextual information has already been compressed and is no longer accessible for conditioning.
Therefore, introducing meta-action embeddings immediately before the history aggregation step, specifically at the \( \mathcal{T}^{\text{env}} \) level, is the only placement that preserves both semantic consistency and sufficiency with the task formulation in Equation~\eqref{eq:task_defintion_c}.

\paragraph{Training}
We freeze the parameters of the foundation model and fine-tune the meta-action prediction module by minimizing the CE loss over the predicted meta-action distribution:
\begin{equation}
    \mathcal{L}_{\text{ma-prediction}} = \text{CE}(\hat{C}, \mathbf{C}_{1:T}).
\end{equation}

\subsection{Meta-Action Injection Module}

To control the trajectory generation process, we inject meta-action information into the foundation model. Specifically, the meta-action sequence \( \mathbf{C}_{1:T} \) is embedded and injected into the history aggregation module, corresponding to \( \text{Attn}_4 \) in Equation~\eqref{eq:cross_atten_time}. This follows the critical injection timing between environment interaction and history aggregation, as discussed previously.

Let the hidden states of the history aggregation module be denoted by:
\begin{equation}
    \mathcal{T}^{(l)} = \text{MHAttn}^{(l-1)}\left(\mathcal{T}^{(l-1)}\right), \quad \text{where } \mathcal{T}^{(0)} = \mathcal{T}^{\text{env}}.
\end{equation}
Here, \( l \in \{1, \ldots, L\} \) is the layer index, \( \text{MHAttn}^{(l-1)} \) denotes the \((l-1)\)-th multi-head attention layer in \( \text{Attn}_4 \), and \( \mathcal{T}^{(l-1)} \) is its input.

To integrate meta-action information, we inject a meta-action embedding at each layer as follows:
\begin{gather}
    \mathcal{T}^{(l)} = \text{MHAttn}^{(l-1)}\left(\mathcal{T}^{(l-1)} + \mathcal{C}_2^{(l-1)}\right),\,
    \text{where } \mathcal{C}_2^{(l-1)} = \text{Embed}_{l+1}(\mathbf{C}_{1:T}).
\end{gather}

After \( L \) layers, the final output is further enhanced by one additional embedding before detokenization:
\begin{equation}
    \mathcal{T}^{\text{ma-history}} = \mathcal{T}^{(L+1)} + \text{Embed}_{L+2}(\mathbf{C}_{1:T}).
\end{equation}

This design introduces a total of \( L+1 \) meta-action embedding layers, where the embedding layers \( \text{Embed}_{l+1} \) are uniquely indexed to distinguish them from \( \text{Embed}_1 \) used in Equation~\eqref{eq:meta-action-shift}.

The meta-action-injected tokens \( \mathcal{T}^{\text{ma-history}} \) are then passed to the FFN layer from the foundation model, as defined in Equation~\eqref{eq:ffn}, to produce the final trajectory prediction.

\paragraph{Training}
 We freeze the foundation model and finetune the meta-action embeddings by:
 \begin{equation}
     \mathcal{L}_{\text{ma-injection}} = \text{CE}(\hat{A},\mathbf{A}_{1:T}).
 \end{equation}

\section{Experiments}
\label{sec:maintext_exp}
This section presents important experimental results and analysis. Complete results and detailed illustrations are available in Appendix \ref{sec:app_supplementary_experiment_results}.
We briefly introduce the used datasets and metrics here, with comprehensive setting illustration and implementation details presented in Appendix \ref{sec:app_exp_details}.

\paragraph{Datasets}
We use the Waymo Motion Dataset v1.2 \citep{sun2020scalability}, generating frame-level meta-action labels using the approach described in Appendix \ref{sec:app_label_generation}.

\paragraph{Metrics}
We use recall, precision, and mean average precision (mAP) to evaluate the decision-following ability, details in Appendix.
We use metrics provided by the \textit{Waymo SimAgents} benchmark to evaluate the quality of generated trajectories, including kinematic, interactive, map-based metrics, and realism score for the overall authentic quality and minADE for the closeness to the ground truth.

\subsection{Comparison to Long-Interval Meta-Actions}
We compare our approach, an autoregressive base model equipped with autoregressive (unit-interval) meta-actions, with conventional methods using long-interval meta-actions in the context of decision-following evaluation. For the regression-based baseline, we adopt a UniAD-style architecture~\citep{hu2023planning} under the formulation of Equation~\eqref{eq:regressive_overall_task}. To represent the autoregressive counterpart with long-interval meta-actions, we reuse our architecture trained under the conventional formulation in Equation~\eqref{eq:autoregressive_condition_task}. We report results in terms of recall, precision, and mAP for the decision following.

\begin{center}
\footnotesize
\begin{tabular}{@{}lllll@{}}
\toprule
MetaAction & Base Model                        & Recall & Precision & mAP   \\ \midrule
LongInterval&Regression             & 0.641  & 0.493     & 0.567 \\
LongInterval& AutoRegressive                & 0.734  & 0.537     & 0.635 \\
AutoRegressive& AutoRegressive & \textbf{0.830}  & \textbf{0.607}     & \textbf{0.718} \\ \bottomrule
\end{tabular}
\end{center}

The regression-based model with long-interval meta-actions performs the worst, while the autoregressive variant improves mAP to 0.635. Our full autoregressive approach, jointly modeling trajectories and meta-actions, further boosts mAP to 0.716 by enhancing task unification. These results support our analysis that autoregressive modeling reduces task complexity, with both frame-level trajectory prediction and unit-interval meta-actions outperforming their long-horizon counterparts.

\begin{figure}[!h]
    \centering
    \includegraphics[width=1.\linewidth]{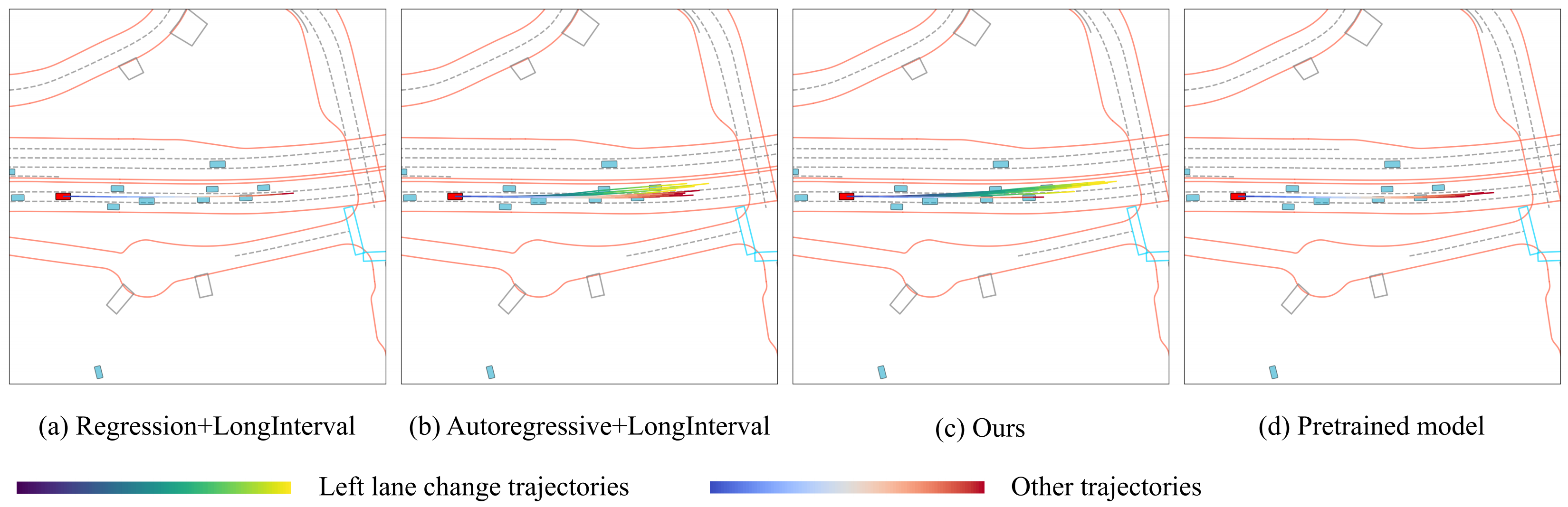}
    \caption{Visual comparison of various models for following decision "Left Lane Change".}
    \label{fig:comparsion}
\end{figure}

For a qualitative evaluation, Figure~\ref{fig:comparsion} visualizes generated trajectories under the decision condition \textbf{Left Lane Change}. The results for Regression+LongInterval, Autoregressive+LongInterval, and our Autoregressive+Autoregressive (Ours) model are shown in Figures~\ref{fig:comparsion}(a), (b), and (c), respectively.
Figure \ref{fig:comparsion}(d) presents results of our foundation model without decision intervention.

We observe that long-interval meta-action models tend to underperform in decision compliance. Specifically, the regression-based model fails to produce lane-changing behavior, instead outputting a trajectory that maintains the current lane. The autoregressive variant shows partial improvement but still yields 4 out of 10 samples that violate the intended decision. This indicates a residual bias toward natural, unconditioned behaviors learned by the foundation model (as shown in Figure \ref{fig:comparsion}(d)). In contrast, our autoregressive meta-action approach generates 9 out of 10 trajectories that conform to the specified decision, demonstrating stronger adherence and controllability.

\subsection{Ablation Studies}
We conduct ablation studies to evaluate the effectiveness of key components in our framework. Specifically, we assess the impact of historical meta-action conditioning (\textbf{HistMA}) by removing the historical meta-action sequence from the model input. To examine the role of staged fine-tuning, we construct an ablation variant (\textbf{Finetuning}) in which only the meta-action-related parameters are trained jointly, without the staged procedure.
\begin{center}
\footnotesize
\begin{tabular}{@{}cccccc@{}}
\toprule
 HistMA & Finetuning                    & Recall & Precision & mAP & Training Time (hour/epoch)  \\ \midrule
 \ding{55}  & \ding{55} & 0.815  & 0.587     & 0.701 & 0.9 \\
 \ding{51}  & \ding{55} & 0.830  & \textbf{0.607}     & \textbf{0.718} & 0.9 \\
 \ding{51}  & \ding{51} & \textbf{0.838}  & 0.593     & 0.716 & \textbf{0.4}\\ \bottomrule
\end{tabular}
\end{center}
As shown, removing historical meta-action input leads to a clear drop in mAP, demonstrating that access to prior decisions enables more accurate localization of decision transitions. This underscores the importance of conditioning on historical meta-actions (see Appendix~\ref{sec:discussion_on_histma} for further analysis). Regarding the training strategy, fine-tuning achieves performance comparable with joint training, while significantly reducing training time, with an approximate 2.25x speed-up compared to training from scratch (assuming integration of new decision types into a pretrained foundation model).

\subsection{Overall Generation Quality}
This experiment assesses the impact of meta-action-related modules on overall trajectory generation quality.
we compare the \textbf{Foundation Model} (FM), which excludes meta-action-related modules, with the \textbf{Controllable Model} (CM), which incorporates these modules.
Results of the official metrics defined by the Waymo SimAgents benchmark are presented below.
\begin{center}
\footnotesize
\begin{tabular}{@{}llllll@{}}
\toprule
                   & REALISM & Kinematic metrics & Interactive metrics & Map-based metrics & minADE \\ \midrule
FM   & 0.7514  & \textbf{0.4531}            & 0.7992               & \textbf{0.8604}            & 1.3394 \\
CM & \textbf{0.7518}  & 0.4530            & \textbf{0.8002}               & \textbf{0.8604}            & \textbf{1.3341} \\ \bottomrule
\end{tabular}
\end{center}
As shown, the two models achieve comparable trajectory generation quality across all metrics, demonstrating that the integration of meta-action components does not degrade performance. Instead, it preserves generation effectiveness while introducing controllability, highlighting the practical plug-and-play capability of our approach.

\subsection{Intent Interpretation Ability and Supplementary Results}
We investigate the intent interpretation ability of our model for explaining generated trajectories with high-level intentions.
These results, along with supplementary results, are available in Appendix \ref{sec:app_supplementary_experiment_results}.

\section{Conclusion}
This paper proposes a unified formulation for controllable trajectory generation based on autoregressive meta-actions. We develop a modular model architecture tailored to this task, which first pretrains a foundation model for generalizable trajectory generation and then fine-tunes additional modules to inject meta-action guidance. This staged design enables flexible and stable integration of decision-following capabilities. Furthermore, we present a labeling strategy for deriving frame-level meta-actions from trajectories, facilitating the adaptation of existing datasets to our new task formulation. Extensive experiments and ablation studies demonstrate the effectiveness of our approach in producing trajectory generation models with strong decision-following performance.

\newpage

\bibliographystyle{plainnat}
\bibliography{ref}

\newpage

\appendix

\section{Related Work}
In this section, we first review the mainstream scene representation approaches used in trajectory generation in Section~A.1. Then, in Section~A.2, we introduce two predominant modeling paradigms: generating long-horizon trajectories in a single step versus generating them incrementally in an autoregressive manner. Finally, in Section~A.3, we summarize and discuss recent efforts on controllable trajectory generation, highlighting methods that enable prediction under high-level guidance.

\subsection{Scene Representations for Trajectory Generation}

Scene representation plays a critical role in trajectory generation, as it determines how contextual information is encoded and utilized by the model. Existing approaches can be broadly categorized into three types: \textit{scene-centric}, \textit{agent-centric}, and \textit{query-centric} representations.

\textbf{Scene-centric representations} encode the entire scene as a global context \citep{chen2024ppad, huang2024gen, sun2023large} and apply standard transformer architectures directly \citep{vaswani2017attention}, without structural modifications. While conceptually straightforward, this design lacks translation invariance—agent positions are interpreted in absolute coordinates\footnote{Here, we define absolute coordinates as those specified in a fixed coordinate system, typically the ego-centric frame at the current timestep.}—which hinders the model's ability to generalize across spatially diverse scenarios. Consequently, such methods often underperform and are rarely adopted in state-of-the-art systems \citep{zhang2024trafficbots}.

\textbf{Agent-centric representations} treat each agent independently and re-center the scene around the agent of interest, effectively transforming all contextual elements (e.g., surrounding agents, map features) into the agent’s local coordinate frame \citep{wang2023prophnet, min2021gatsbi, feng2023macformer}. This strategy aligns all predictions to a shared reference system, significantly reducing task complexity and enabling higher performance. However, it requires repeated inference for each agent \citep{shi2022motion}, resulting in substantial computational overhead and longer inference times.

\textbf{Query-centric representations} modify the transformer architecture by introducing an $N^2$-sized pairwise relative position embedding (RPE) matrix to explicitly encode spatial relationships among scene elements \citep{zhou2023query, zhou2023qcnext}. Leveraging relative spatial features to model inter-agent dependencies, thereby enhancing permutation and translation invariance. 
Due to their balance between accuracy and flexibility, query-centric methods have gained widespread adoption in recent transformer-based motion generation frameworks.
More recently, \citet{zhao2025drope} extended this direction by introducing rotary position embedding (RoPE) to the domain, proposing a directional variant to capture heading relationships. This significantly reduces the memory complexity by a factor of $N$, the number of agents in the scene, thereby accelerating training while maintaining strong performance. Motivated by its efficiency, we adopt RoPE in our foundation model as well.

\subsection{Single-Step v.s. Autoregressive Trajectory Generation}
In terms of task formulation, trajectory generation tasks are commonly formulated in two paradigms: \textit{single-step} prediction and \textit{autoregressive} (step-by-step) prediction.

\textbf{Single-step prediction.}  
An intuitive task formulation \citep{hu2023planning, wang2025omnidrive} is to generate a complete future trajectory in a single step, conditioned on past states. To address the inherent multimodality of future motion, many approaches predicted \( k \) future trajectories \citep{zhang2024sparsead, Narayanan_2021_CVPR, zhang2024demo} and adopted a \textit{winner-take-all} strategy, where the loss was computed only for the closest predicted trajectory to the ground truth. This effectively assumes a predictive distribution composed of \( k \) Gaussian components and learns only the means by regressing each mode.

Alternatively, some methods employ Gaussian Mixture Model (GMM) loss functions to directly fit a multimodal distribution by optimizing both the means and variances of each Gaussian component \citep{zhou2023query, zhuang2024streammotp, shi2022motion, shi2024mtr++}, without relying on winner-take-all selection. However, both approaches make strong assumptions about the shape of the future distribution. In complex real-world scenarios, such fixed priors may poorly match the true underlying distribution, leading to distributional shifts and potentially unsafe predictions \citep{zhao2024kigras}.

To address this limitation, methods such as VADv2 \citep{chen2024vadv2} discretize the trajectory space into a fixed-size vocabulary and model trajectory generation as a classification task using cross-entropy loss \citep{Goodfellow-et-al-2016}. This avoids explicit assumptions on the distribution, but introduces a trade-off: larger vocabularies improve fidelity at the cost of training difficulty, while smaller vocabularies suffer from high quantization error \citep{lin2025revisit}. More recently, diffusion-based models \citep{diffusiondrive, zheng2025diffusionbased} have been introduced to model the full predictive distribution by iteratively denoising trajectory samples, thus eliminating the need for strong priors. Despite their flexibility, diffusion models struggle to expose explicit probability distributions \citep{ho2020denoising, song2021scorebased}, making them incompatible with planning methods that require tractable likelihoods or search orderings (e.g., MCTS \citep{silver2016mastering, silver2010monte}). Moreover, the sampling difficulty increases sharply when reducing the number of diffusion steps, limiting real-time applicability.

\textbf{Autoregressive prediction.}  
Recently, trajectory generation is decomposed into a step-by-step prediction problem using autoregressive modeling like language models \citep{achiam2023gpt}. Rather than predicting the entire trajectory at once, these models generate one frame at a time and condition future predictions on previously generated outputs \citep{zhou2024behaviorgpt, hu2024solving}. Autoregressive models often assume temporal stationarity, enabling the same set of parameters to model the per-step predictive distribution, which greatly reduces learning complexity. This formulation has been shown to facilitate better multi-agent interaction modeling and scene-aware behavior.

However, most existing autoregressive methods still operate in trajectory (position) space, which can be redundant and noisy. To address this, KiGRAS \citep{zhao2024kigras} introduces a kinematic-guided representation by transforming observed trajectories into control actions, enabling the model to learn underlying driving behavior patterns directly in the control space. This further simplifies the learning task and improves generalizability. Our work builds upon this task formulation.

\subsection{Controllable Trajectory Generation}
Controllable trajectory generation is a crucial capability for autonomous driving, particularly in planning systems. To enable goal-directed behaviors, such as adhering to a navigation route or executing high-level driving decisions, a trajectory generator must respond effectively to specified commands or constraints \citep{afshar2024pbp}. However, despite its importance, controllability has received limited attention in the trajectory prediction literature, largely due to the absence of a clear and unified task formulation in the community.

Early works \citep{hu2023planning, jiang2023vad} introduced simple controllability via pre-defined high-level commands, often referred to as \textit{ego commands}. These commands typically fall into three coarse categories: \textit{go straight}, \textit{turn left}, and \textit{turn right}. They are usually derived from the lateral offset of the ego vehicle over a short future horizon (e.g., 3 seconds), and are applied uniformly to guide trajectory generation. However, these methods lack explicit interaction with the road topology (e.g., lane structure), making them unreliable in scenarios such as lane changes. Moreover, assigning a single command to control a long trajectory segment fails to handle meta-action transitions effectively, particularly near decision boundaries.

Another line of work introduces controllability by conditioning on the endpoint \citep{zhao2021tnt, lin2024eda} of the trajectory over a long horizon (typically 8 seconds). These methods aim to guide the trajectory toward a specific final position, but such endpoints often do not reflect the intermediate decision process or semantics of the maneuver, leading to weak interpretability and limited reactivity.

Both types of approaches can be abstractly described as conditioning the model on a high-level signal \( c \) (e.g., a command or endpoint). More recent methods go further by not only conditioning on \( c \), but also estimating its distribution based on historical and current observations—thereby allowing the model to reason about likely high-level decisions \citep{zhang2025carplanner, hwang2024emma}. Nevertheless, these models still rely on long-horizon conditioning, which introduces alignment issues during meta-action transitions. As illustrated in Figure \ref{fig:ma_example} in the main texts, this often results in inconsistent behavior around decision boundaries.

To address the gap in unified task formulation, we propose an \textit{autoregressive meta-action-based framework}, where meta-actions are predicted and applied at each frame instead of a long interval. This formulation ensures strict temporal alignment between high-level decisions and low-level motion, significantly simplifies task complexity, and enables strong controllability.

\section{Model Details}
\label{sec:app_model_details}

In this section, we provide implementation details of our architecture. We first describe our agent and map encoders, which transform traffic scene elements into token representations. We then introduce the RoPE-based attention and the derivation of kinematic control action labels.

\subsection{Agent Encoder}
In the agent encoder, we process all dynamic entities in the scene, including vehicles, pedestrians, cyclists, and others. For each agent, we extract its physical attributes from the dataset, such as dimensions (length, width, height) and frame-wise dynamic properties (e.g., speed). At each timestep, we encode these scalar values—typically with inherent magnitude relationships—using a multi-layer perceptron (MLP) to obtain a feature vector. This is then added to a learned agent-type embedding to form the final token representation \( \mathcal{T} \).  
Note that we do not embed any spatial position information into the agent tokens at this stage.

\subsection{Map Encoder}
For each map element represented as a polyline composed of a sequence of spatial points, we construct a local coordinate system by setting the midpoint of the polyline as the origin, and aligning the positive \( x \)-axis with the direction from the midpoint to its immediate successor. The polyline is then transformed into this local frame, which effectively removes the influence of spatial positions from the map representation.

To capture the shape and semantics of each polyline, we apply a subgraph \citep{Gao_2020_CVPR} encoder to its point sequence. We embed categorical information—such as the polyline type (e.g., lane centerline, stop line) and traffic light status—using an embedding layer. Additional attributes (e.g., distance to stop line, speed limits) are encoded via an MLP. These components are summed to produce the final map token \( \mathcal{M} \).

\subsection{Rotary Position Embedding for Spatial Attention}
This part introduces the rotary position embedding (RoPE)-based attention used in the model, which is referred to as the following formulation in the main texts:
\begin{equation}
    \text{Attn}^{\text{RoPE}}(\mathbf{Q},\mathbf{K},\mathbf{V}),
\end{equation}
where $\mathbf{Q}\in \mathbb{R}^{q\times d},\mathbf{K}\in \mathbb{R}^{k\times d},\mathbf{V}\in \mathbb{R}^{k\times d_v}$ represent the query, key, and value tokens.
Denote $\text{Attn}$ as a standard attention operation, then the RoPE-based attention is represented as:
\begin{equation}
    \text{Attn}^{\text{RoPE}}(\mathbf{Q},\mathbf{K},\mathbf{V}) \equiv \text{Attn}(\text{RoPE}(\mathbf{Q},\mathbf{p}^{\mathbf{Q}}),\text{RoPE}(\mathbf{K},\mathbf{p}^{\mathbf{K}}),\mathbf{V}).
    \label{eq:attn(RoPE)}
\end{equation}
Here, $\mathbf{p}^{\mathbf{Q}} \in \mathbb{R}^{q\times 1},\mathbf{p}^{\mathbf{K}}\in \mathbb{R}^{k \times 1}$ are the positions associated with the corresponding query and key tokens, repetitively, where $\mathbf{p}^{\mathbf{Q}}_{i}$ is the position of token $\mathbf{Q}_i$.

The RoPE transformation in Equation~\eqref{eq:attn(RoPE)} is formally defined as:
\begin{gather}
    \text{RoPE}(\mathbf{Q}, \mathbf{p}) = \left[ R(\mathbf{Q}_{i,:}, \mathbf{p}_i) \right]_{i=1}^q, \quad \text{where } R: \mathbb{R}^d \times \mathbb{R} \rightarrow \mathbb{R}^d, \\
    R(Q, p) = \left[ r_i(Q_{2i-1:2i}, p) \right]_{i=1}^{d/2}\in \mathbb{R}^d, \quad \text{where } r_i: \mathbb{R}^2 \times \mathbb{R} \rightarrow \mathbb{R}^2, \\
    r_i(q, p) = 
    \begin{bmatrix}
        \cos(p \theta_i) & -\sin(p \theta_i) \\
        \sin(p \theta_i) & \cos(p \theta_i)
    \end{bmatrix} q^\top, \quad \text{with } \theta_i = 100^{-2i/d}.
    \label{eq:r_i}
\end{gather}

In essence, each token is extended to an even-dimensional vector \( d \), and a rotary transformation is independently applied to each 2D slice \( Q_{2i-1:2i} \) using a frequency-scaled angle \( p\theta_i \). The rotated 2D components are then concatenated to construct the final position-encoded token. In this manner, the positional information \( p \) is embedded into the token representation through structured rotation.

Thanks to the rotational property \( r(\alpha) \cdot r(\beta)^\top = r(\alpha - \beta) \), the resulting attention scores depend only on relative positions between tokens. This implicitly encodes relative positional information without the need for explicit subtraction or additional computation. A formal proof of this result is provided in the original RoFormer paper~\citep{su2024roformer}.

In the context of trajectory generation, the spatial information that describes the position of an object includes its spatial coordinates $(x,y)$, and heading direction $\theta$.

For the spatial position $(x,y)$, we encode this 2D position vector by:
\begin{equation}
    R^{\text{2D}}(Q,x,y) = [r_i(Q_{4i-3:4i-2},x),r_i(Q_{4i-1:4i},y)]_{i=1}^{d/4} \in \mathbb{R} ^d,
\end{equation}
using the rotary function $r_i$ defined by Equation \eqref{eq:r_i}.
Here, the tokens are extended to a dimension $d$ that is a multiple of $4$ to simultaneously encode coordinates $(x,y)$.

For the heading direction $h$, we adopt the directional RoPE proposed by \citet{zhao2025drope}:
\begin{equation}
\hat{r}_i(q, p) =
\begin{bmatrix}
\cos(p) & -\sin(p) \\
\sin(p) & \cos(p)
\end{bmatrix} q^\top,
\end{equation}
where the original frequency-modulated term $\theta_i$ in the rotary function $r_i$ (defined in Equation~\eqref{eq:r_i}) is replaced by a constant scalar $1$. 
The other operations remain the same as RoPE.
This adjustment ensures that the embedding retains the inherent $2\pi$-periodicity of directional angles.

We adopt a head-by-head design to integrate relative spatial coordinates and heading directions in a mixed manner, following the approach introduced by \citet[Section IV-A]{zhao2025drope}.

\subsection{Kinematic Control Action Inference from Trajectories}

This part introduces how to generate the kinematic control actions $\mathbf{A}_{1:T}$ from trajectory states $\mathbf{X}_{1:T}$ as the training labels.
We adopt the Constant Turn Rate and Acceleration (CTRA) model~\citep{stone2012kinematic} to characterize the kinematic control action. The agent state is represented as \( \mathbf{x}_t = (x, y, \theta, v) \), comprising the spatial pose \((x, y, \theta)\) and velocity \(v\). The corresponding control action is defined as \( \mathbf{a}_t = (\text{acc}, \dot{\theta}) \), where \(\text{acc}\) denotes acceleration and \( \dot{\theta} \) denotes the yaw rate.
Given the state transition sequence \( \mathbf{X}_{t:t+1} \), we infer the control sequence \( \mathbf{A}_{0:T-1} = \{ \mathbf{A}_t \} \) using a rolling-horizon optimization strategy inspired by Model Predictive Control (MPC)~\citep{garcia1989model}.

At each time step \( t \), a continuous control sequence is optimized over a finite prediction horizon of length \( k \), aiming to minimize the deviation between the forward-simulated trajectory and the ground-truth trajectory. The optimization problem is formulated as:

\begin{equation}
\label{eq:opt_mpc_ctra}
\begin{aligned}
    & \min_{\mathbf{a}_{t:t+k-1}} \sum_{\tau=t}^{t+k-1} \left\| \mathbf{x}_{\tau+1}^{\text{ctl}} - \mathbf{x}_{\tau+1} \right\| \\
    & \text{subject to} \quad \mathbf{x}_{\tau+1}^{\text{ctl}} = \mathcal{K}(\mathbf{x}_\tau^{\text{ctl}}, \mathbf{a}_\tau)
\end{aligned}
\end{equation}

Here, \( \mathcal{K} \) denotes the CTRA \citep{werling2010optimal} transition function. After solving for the continuous control sequence \( \{\mathbf{a}_t', \ldots, \mathbf{a}_{t+k-1}'\} \), we project the first optimized control \( \mathbf{a}_t' \) to its closest discrete bin in the action space, resulting in the discrete action label \( \mathbf{a}_t \). The corresponding next state \( \mathbf{x}_{t+1}^{\text{ctl}} \) is then computed and used as the initial condition for the subsequent rolling window.

This procedure is repeated iteratively across the entire trajectory, enabling frame-level action label extraction that adheres to the CTRA dynamics.

\section{Label Generation of Frame-Level Meta-Actions}
\label{sec:app_label_generation}

This section introduces how to automatically infer frame-level meta-actions from observed vehicle trajectories and lane geometry, which involves (1) Temporal Behavior Decomposition: Segmenting continuous trajectories into discrete action units (e.g., lane keeping, lane changes, turns) at each timestep.
(2) Context-Aware Labeling: Using both kinematic features (e.g., curvature, lateral displacement) and road topology (e.g., lane connectivity, turn restrictions) to resolve ambiguous motion patterns.
(3) Frame-Level Annotation: Assigning a meta-action label (e.g., keep lane, left lane change) to each trajectory point, enabling fine-grained behavior analysis.

\subsection{Classification Criteria of Frame-level Meta-Actions}

We define a set of frame-level meta-action labels $\mathcal{M}$ to describe the semantic driving behavior of each trajectory point $p_t$ within a scene. These labels are derived from low-level kinematic cues and high-level road topology for interpretable motion behavior, which includes in our setting: 
$$
\mathcal{M} = \left\{ \text{stationary}, \text{keep lane}, \text{lane change}, \text{turn}, \text{U-turn} \right\}
$$

\textbf{Stationary}. The \textit{stationary} state refers to situations where the vehicle comes to a complete stop, typically due to external factors such as traffic lights, stop signs, or road congestion. It is characterized by near-zero velocity and minimal spatial displacement over a short temporal window. Each trajectory point is represented as \( p_i = (x_i, y_i, v_i) \), where \( (x_i, y_i) \) denotes the 2D position of the vehicle and \( v_i \) its instantaneous speed at time step \( i \). Given a local window \( W_t = \left\{ p_{t-n}, \dots, p_{t+n} \right\} \), we compute the average velocity and displacement as follows:
\begin{equation}
\bar{v}_t = \frac{1}{2n+1} \sum_{i=t-n}^{t+n} v_i
\end{equation}
\begin{equation}
\bar{s}_t = \frac{1}{2n} \sum_{i=t-n}^{t+n-1} \left\| p_{i+1}^{(xy)} - p_i^{(xy)} \right\|
\end{equation}
Here, \( p_i^{(xy)} = (x_i, y_i) \) extracts the spatial coordinates from \( p_i \), and \( \left\| \cdot \right\| \) denotes the Euclidean norm.

The stationary condition holds if:

\begin{equation}
\bar{v}_t < \varepsilon _v, \mathrm{or}\ \bar s_t < \varepsilon _s
\end{equation}

Typical thresholds: $\varepsilon _v=0.3 $m/s, $\varepsilon _s=0.1$m

\textbf{Keep Lane}. The keep-lane behavior denotes stable longitudinal motion along a lane’s centerline with minimal curvature and lateral deviation. It reflects straight-line driving on a consistent heading.

\begin{equation}
\left | \kappa_t  \right | < \varepsilon _{\kappa}, \mathrm{or} \ \left | d_{y,t} \right | < \epsilon _{d} , \mathrm{or}\ \Delta l_t=0.
\end{equation}

Where:
\begin{itemize}
    \item $\kappa_t$: curvature of the trajectory at time $t$ estimated via cubic fitting,
    \item $d_{y,t}$: lateral offset in Frenet frame,
    \item $\Delta l_t$: number of distinct lane IDs in $W_t$.
\end{itemize}
Typical thresholds:
$\varepsilon _{\kappa }=0.015 \mathrm{m^{-1} } $, $\epsilon_{d}=0.3 \mathrm{m}$

\textbf{Lane Change}.   We define a lane change as a lateral maneuver in which a vehicle transitions from its current lane to an adjacent target lane, typically crossing one or more lane boundaries. Typically, lane changes: (1) maintain forward longitudinal motion (small yaw change), (2)exhibit lateral deviation across lane boundaries, (3)  complete over a short temporal horizon (usually 3–5 seconds), (4) and maintain continuity in speed and heading direction.
Lane change behavior involves left and right maneuvers
 based on the sign of lateral offset from the lane centerline.

To formally identify lane change behavior according to the analyzed characteristics before, we adopt the following criteria:
\begin{equation}
|d_{y,t}| > d_{\min}, \quad \Delta l_t \ge 1, \quad \exists\, i, j \in W_t: \kappa_i \cdot \kappa_j < 0.
\end{equation}

Where:
\begin{itemize}
    \item $d_{y,t}$: signed lateral offset from the lane centerline at time $t$, where $d_{y,t} > 0$ indicates a left lane change, and $d_{y,t} < 0$ indicates a right lane change.
    \item $d_{\min} = 1.75\,\text{m}$: minimum lateral displacement to trigger lane change,
    \item $\Delta l_t$: number of lane binding changes in window $W_t$,
    \item $\kappa_i$: curvature value at frame $i$,
    \item The sign change $\kappa_i \cdot \kappa_j < 0$ in curvature reflects the characteristic S-shaped motion.
\end{itemize}
\textbf{Turning}. The turning behavior denotes sustained lateral curvature in a single direction. A vehicle in a turn maintains spatial alignment with a designated turning path while exhibiting a consistent heading angle change. Unlike lane changes, turns do not involve lane ID transitions but are characterized by moderate and unidirectional curvature. 
Similarly, turning behavior includes left turns and right turns based on the direction of curvature of the trajectory. 

To formally identify turning behavior, we adopt the following criteria:
\begin{equation}
|\kappa_t| \in [\kappa_{\text{min}},\ \kappa_{\text{max}}],\quad \Delta \theta_t \geq \theta_{\text{min}},\quad \forall i \in W_t: \operatorname{sign}(\kappa_i) = \text{const}.
\end{equation}

\noindent
Where:
\begin{itemize}
\item $\kappa_t$: the curvature of the trajectory at time $t$, where $\kappa_t < 0$ indicates a left turn and $\kappa_t > 0$ indicates a right turn.
    \item $\Delta \theta_t$: total heading angle change over the window $W_t$,
    \item $\operatorname{sign}(\kappa_i) = \text{const}$: curvature remains unidirectional (either all left or all right) within the window.
\end{itemize}

The thresholds we use are presented below:
\[
\kappa_{\text{min}} = 0.015\, \text{m}^{-1},\quad \kappa_{\text{max}} = 0.25\, \text{m}^{-1},\quad \theta_{\text{min}} = 15^\circ.
\]

\textbf{U-turn}. U-turns are a special case of turning behavior characterized by extreme curvature and a complete reversal in heading direction. Unlike regular turns, U-turns require the vehicle to traverse a median opening in the road topology while maintaining unidirectional curvature throughout the maneuver.

To formally identify left and right U-turn behaviors, we extend the original turn criteria with enhanced thresholds:

\begin{equation}
\begin{cases}
\kappa_t \in (+\kappa_{\text{max}}, +\infty) & \text{(Right U-turn)} \\
\kappa_t \in (-\infty, -\kappa_{\text{max}}) & \text{(Left U-turn)}
\end{cases}, \, 
\Delta\theta_t \in [180^\circ - \alpha, 180^\circ + \alpha], \, \forall i \in W_t: \operatorname{sign}(\kappa_i) = \text{const}.
\end{equation}

The thresholds we use are presented below:

\begin{equation}
\kappa_{\text{min}} = 0.015\,\text{m}^{-1}, \quad \kappa_{\text{max}} = 0.25\,\text{m}^{-1}, \quad \alpha = 15^\circ
\end{equation}

\section{Experimental Details}
\label{sec:app_exp_details}

\subsection{Benchmark Construction}
To evaluate the model’s ability to follow high-level decisions, we mine the validation set to identify agents executing distinct meta-actions. We construct a curated subset of 661 scenes, each selected through rule-based filtering and manual verification to ensure the presence of specific meta-action behaviors. Since the dataset contains many agents per scene, we focus our analysis on the ego vehicle's meta-actions.

Due to the limited 1-second trajectory length available in the test set—which is insufficient for reliable behavior inference—all experiments are conducted on the validation set. This study focuses on five representative meta-actions: Keep Lane, Left Lane Change, Right Lane Change, Turn Left, and Turn Right.

\subsection{Metrics Design}
Inspired by the use of precision and recall metrics in object detection, we adapt these measures to suit the characteristics of the trajectory-following task under meta-action guidance. Specifically, for a given meta-action, we sample \( N \) candidate trajectories. If all sampled trajectories satisfy the behavioral constraints associated with the target meta-action, the corresponding test scenario is counted as a true positive (TP). The recall is then computed as the proportion of TP scenarios among all evaluation samples.

To assess precision, we incorporate a driving rationality criterion: among the \( N \) sampled trajectories, we identify a subset \( N_p \) that conforms to realistic and safe driving behaviors. The per-scenario precision is defined as \( N_p / N \), and the global precision is obtained by averaging this ratio across all test scenarios.

This evaluation framework ensures that behavior patterns violating normal driving logic—such as consecutive unnecessary lane changes—are penalized as unreasonable samples. As a result, the metrics jointly reflect both the model's ability to follow meta-actions and the plausibility of its driving behavior in realistic scenarios.

\subsection{Implementation Details}
We fix the embedding dimension of all tokens to 64. The environment fusion module is repeated three times to sufficiently capture scene-level interactions. Both the action prediction module and the meta-action prediction module consist of three transformer layers for temporal aggregation.

We train the model using the Adan optimizer~\citep{xie2024adan} with a fixed learning rate of 1e-2 and no learning rate decay schedule. The batch size is set to 64, meaning that each training iteration processes 64 complete driving scenes in parallel. Since each scene typically contains a large number of agents, and we estimate control distributions for each agent at every timestep, the effective batch size in terms of learning signal is considerably larger—approximately equal to the batch size multiplied by the average number of agents per scene and the temporal window length.

In our setting, the temporal window spans 9 seconds and includes 1 second of historical data, resulting in dense, agent-level supervision. To accommodate this large effective batch size and ensure efficient convergence, we adopt a relatively high learning rate. All models are trained on 8 NVIDIA A800/H20 GPUs.

\section{Supplementary Experimental Results}
\label{sec:app_supplementary_experiment_results}
This section first presents evaluation results demonstrating the good intent interpretation ability of our model, followed by an in-depth analysis of the necessity of incorporating historical meta-actions. Lastly, we provide additional visualizations and supporting analysis to further illustrate and clarify the model’s behavior beyond the results discussed in the main paper.

\subsection{Evaluation on Intent Interpretation Ability}
This section evaluates the performance of our model as an interpretable tool for revealing the high-level intentions behind generated trajectories. This interpretability is enabled by our meta-action-related modules, which automatically infer the next-step meta-action and subsequently guide trajectory generation through meta-action injection. In essence, beyond generating motion trajectories, our model provides semantic explanations of agent behavior over time. This capability is particularly valuable in safety-critical applications such as human-in-the-loop planning and behavior monitoring, where understanding the underlying intent of actions is crucial. We demonstrate this interpretive ability through a series of qualitative visualizations.

\begin{figure}[!h]
\centering
\includegraphics[width=1.\linewidth]{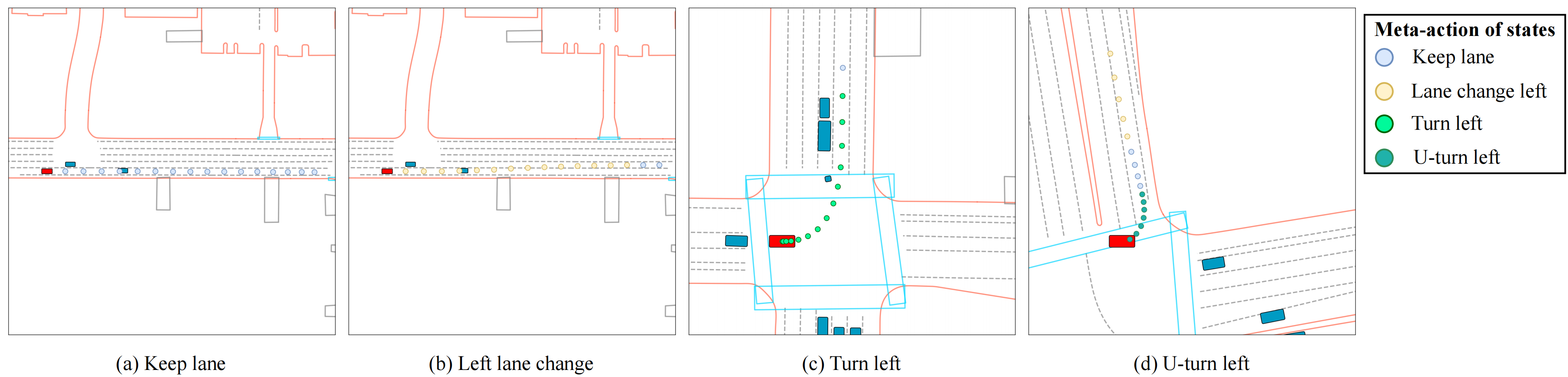}
\caption{Frame-by-frame visualization of predicted meta-actions and resulting ego trajectories. Colored dots represent the ego vehicle’s state and corresponding meta-action at each timestep. Different subplots illustrate behaviors under autoregressively sampled meta-actions or specific injected meta-actions such as Left Lane Change and Turn Left.}
\label{fig:ma_demo}
\end{figure}

Figure~\ref{fig:ma_demo} presents visualizations of sampled meta-actions and the resulting trajectory rollouts from our model. The ego vehicle is highlighted in red. In Figure~\ref{fig:ma_demo}(a) and (d), the meta-actions at each timestep are sampled from the predicted meta-action probability distributions. In contrast, Figure~\ref{fig:ma_demo}(b) and (c) illustrate controlled interventions by manually injecting specific meta-actions at the current frame—left lane change in (b) and turn left in (c). For each case, we visualize the closed-loop simulation trajectory over 8 seconds, where the ego vehicle’s state and meta-action at each timestep are indicated by colored markers.

In Figure~\ref{fig:ma_demo}(a), the ego vehicle maintains a straight path with the predicted meta-actions remaining in the keep-lane state. In Figure~\ref{fig:ma_demo}(b), injecting a left lane change meta-action at the initial frame triggers the vehicle to gradually shift lanes. Upon completing the maneuver, the model autonomously switches the meta-action back to keep lane, demonstrating seamless decision transitions. Figure~\ref{fig:ma_demo}(c) shows a similar transition from a left-turn meta-action to a keep-lane state once the turn is completed. Finally, Figure~\ref{fig:ma_demo}(d) illustrates a more complex behavior sequence: the ego vehicle performs a left U-turn, briefly maintains a keep-lane state, and then initiates a left lane change.

These results highlight the model’s ability to execute fine-grained, interpretable high-level behaviors and adaptively switch between them. We provide additional per-frame visualizations of the meta-action probability distributions in Appendix~\ref{sec:supp_vis}.

\subsection{Analysis of the Integration of Historical Meta-Actions}
\label{sec:discussion_on_histma}
In this section, we further analyze the impact of historical meta-action conditioning (HistMA) on different decision types. We focus on five representative behaviors: Keep Lane, Left Lane Change, Right Lane Change, Turn Left, and Turn Right. The evaluation results are summarized in Table~\ref{wohistma}.

\begin{table}[h!]
\centering
\caption{Comparison of per-category mAP with and without historical meta-action conditioning (HistMA).}
\begin{tabular}{@{}llll@{}}
\toprule
Meta-Action Type  & mAP w/ HistMA & mAP w/o HistMA & $\Delta$ mAP \\ \midrule
Keep Lane         & 0.794         & 0.789          & -0.005       \\
Left Lane Change  & 0.489         & 0.441          & -0.048       \\
Right Lane Change & 0.781         & 0.656          & \textbf{-0.125}       \\
Turn Left         & \textbf{0.809}         & \textbf{0.791}          & -0.018       \\
Turn Right        & 0.652         & 0.683          & +0.031       \\ \bottomrule
\end{tabular}
\label{wohistma}
\end{table}

From Table~\ref{wohistma}, we observe that the influence of HistMA varies notably across different decision categories. Lane change behaviors, especially Right Lane Change, exhibit the most significant degradation when historical meta-action input is removed (a 12.5\% drop in mAP), followed by Left Lane Change (4.8\%). This indicates that such decisions are more sensitive to meta-action transitions, as they typically unfold over multiple frames and require temporal tracking of driver intent. Incorporating historical meta-actions helps the model capture this temporal evolution, improving its ability to accurately detect when a lane change is initiated, underway, or completed.

In contrast, tasks like Keep Lane and Turn Left show minimal performance drop, suggesting that these behaviors are either more temporally stationary or sufficiently discernible from current dynamics and scene context alone.

Interestingly, Turn Right shows a slight improvement in the absence of historical conditioning. This could be due to label noise or misalignment in predicted historical meta-actions for this category, or the possibility that right-turn scenarios in the dataset are structurally simpler and more directly inferable from immediate observations.

\begin{figure}[!h]
\centering
\includegraphics[width=1.\linewidth]{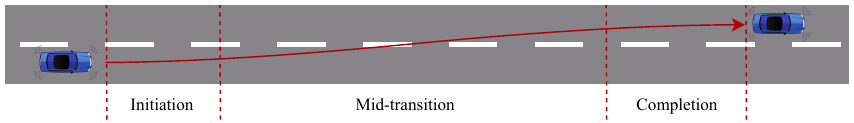}
\caption{Illustration of the temporal stages of a lane change maneuver. The process is divided into three phases: initiation (the vehicle begins lateral deviation), mid-transition (the vehicle straddles the lane boundary), and completion (the vehicle stabilizes within the target lane). Accurate identification of these stages is essential for consistent meta-action prediction and interpretable decision modeling.}
\label{fig:lanechange_stages}
\end{figure}

To further illustrate this point, Figure~\ref{fig:lanechange_stages} decomposes a typical lane change maneuver into three temporal phases: initiation, mid-transition, and completion. Without access to historical decision context, the model must rely solely on a short observation window (e.g., 1 second), making it difficult to distinguish whether the controlled vehicle is just beginning to change lanes or is already completing the maneuver. This temporal ambiguity reinforces the value of historical meta-actions in improving stage-wise reasoning for complex decisions.

This benefit can also be understood from a sequence modeling perspective. Suppose the recent trajectory dynamics at time \( t \) appear similar during early and late stages of a lane change. In this case, the model receives nearly identical input states \( \mathbf{S}_{1:t} \), making it challenging to disambiguate behavior. However, with access to the historical meta-action sequence \( \mathbf{C}_{1:t-1} \), the model can exploit temporal decision continuity to infer context. For instance, consider a transition from a Keep Lane (KL) behavior to a Left Lane Change (LLC). The historical meta-action sequences for the early and late stages differ:
\begin{itemize}
    \item A history of \{KL, KL, KL, LLC\} suggests a newly initiated lane change.
    \item A history of \{LLC, LLC, LLC, LLC\} indicates a continuing or nearly completed maneuver.
\end{itemize}
Even if the state inputs are similar in both cases, the distinct historical decision patterns allow the model to localize behavior phases more accurately. This reinforces the importance of explicitly modeling decision history for temporally extended or multi-phase behaviors.

We hypothesize that the variation in performance across different meta-action types may stem from label imbalance within the dataset.

\subsection{Supplementary Visualized Cases}
\label{sec:supp_vis}
In this section, we provide additional visualizations of meta-action probability distributions and their effects on the predicted control signals (acceleration and yaw rate) of the ego vehicle under different meta-action injection settings and driving scenarios. For non-ego agents, we use the foundation model without meta-action conditioning for inference. We present the full 8-second \textbf{closed-loop simulation} results generated by our model to demonstrate its decision-following behavior over time.

To the right of each BEV (bird’s-eye view) snapshot, we visualize three distributions: meta-action probability, acceleration distribution, and yaw rate distribution. For meta-actions, we focus on seven key behaviors: Keep Lane (KL), Left Lane Change (LLC), Right Lane Change (RLC), Turn Left (TL), Right Turn (RT), Left U-turn (LU), and Right U-turn (RU). In the acceleration distribution, the leftmost bin indicates maximum deceleration, the center corresponds to constant velocity, and the rightmost represents maximum acceleration. Similarly, in the yaw rate distribution, the leftmost bin denotes maximum left turn rate, the center indicates straight driving, and the rightmost represents maximum right turn rate.

\begin{figure}[!h]
\centering
\includegraphics[width=0.71\linewidth]{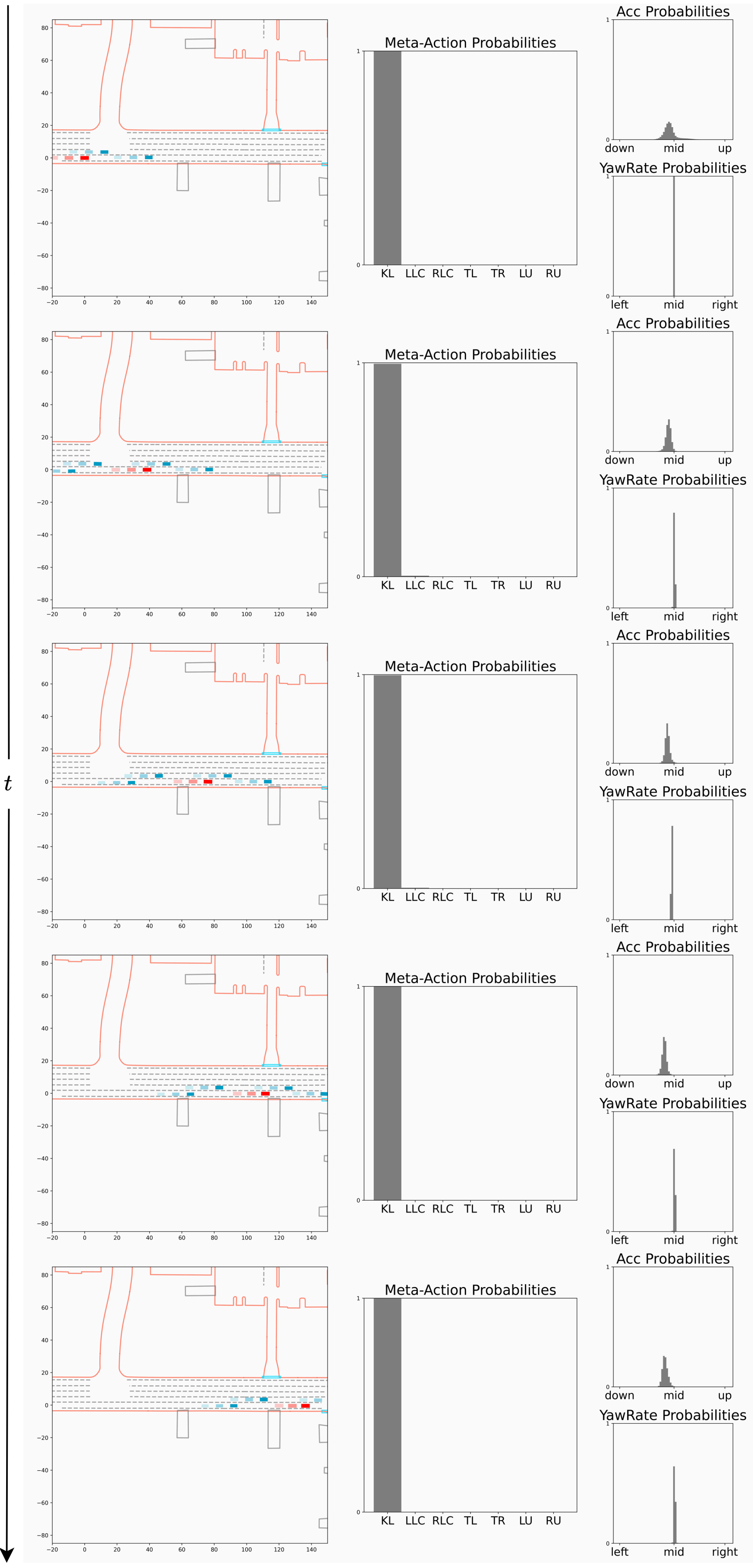}
\caption{Inference result without manually injected meta-actions. The ego vehicle performs standard lane-keeping behavior as the model samples meta-actions and control signals purely from its learned policy.}
\label{fig:kl_vis_woma}
\end{figure}

\begin{figure}[!h]
\centering
\includegraphics[width=0.71\linewidth]{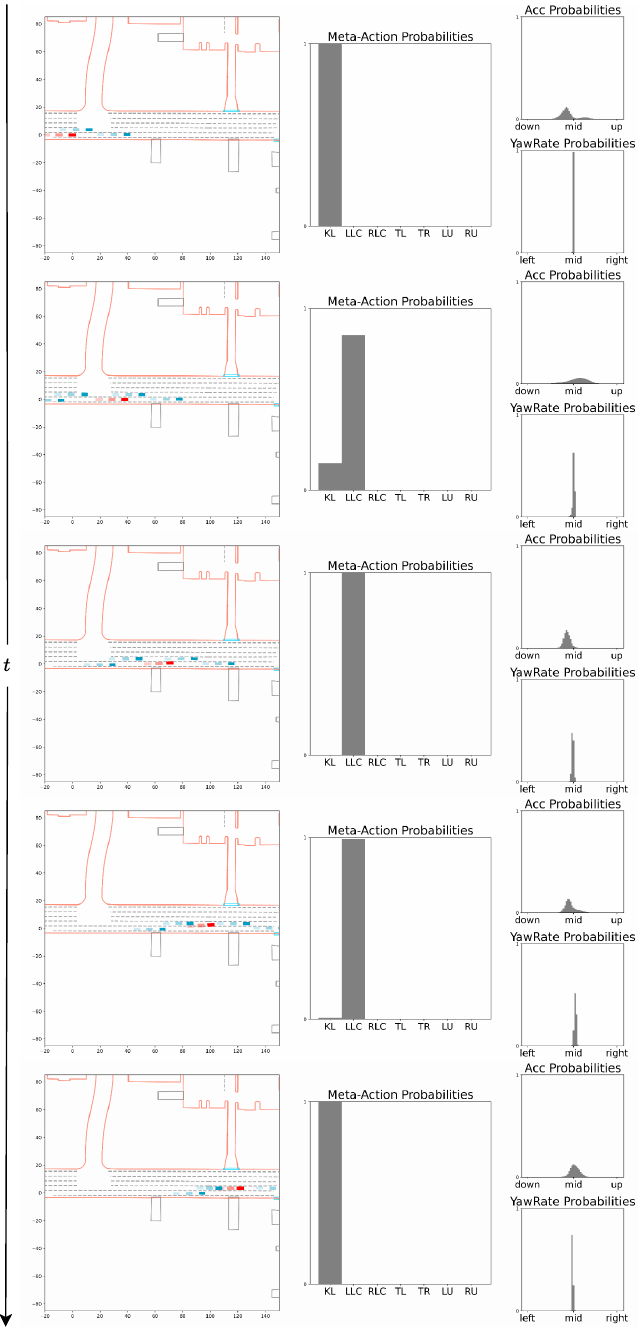}
\caption{Inference result with a manually injected Left Lane Change meta-action at the current frame. The ego vehicle adapts its trajectory accordingly, demonstrating the model’s controllability and responsiveness to external decision guidance.}
\label{fig:llc_vis_ma_in}
\end{figure}

We first present the standard inference results without manually injected meta-actions in Figure~\ref{fig:kl_vis_woma}. Subsequently, Figure~\ref{fig:llc_vis_ma_in} illustrates the model’s inference behavior when a Left Lane Change meta-action is explicitly injected at the current frame.

\newpage

\newpage

As shown, when the LLC meta-action is injected, the ego vehicle gradually initiates and completes a left lane change, after which the predicted meta-action transitions back to keep lane, demonstrating adaptive behavior in line with the intended high-level command.

\begin{figure}[!h]
\centering
\includegraphics[width=0.71\linewidth]{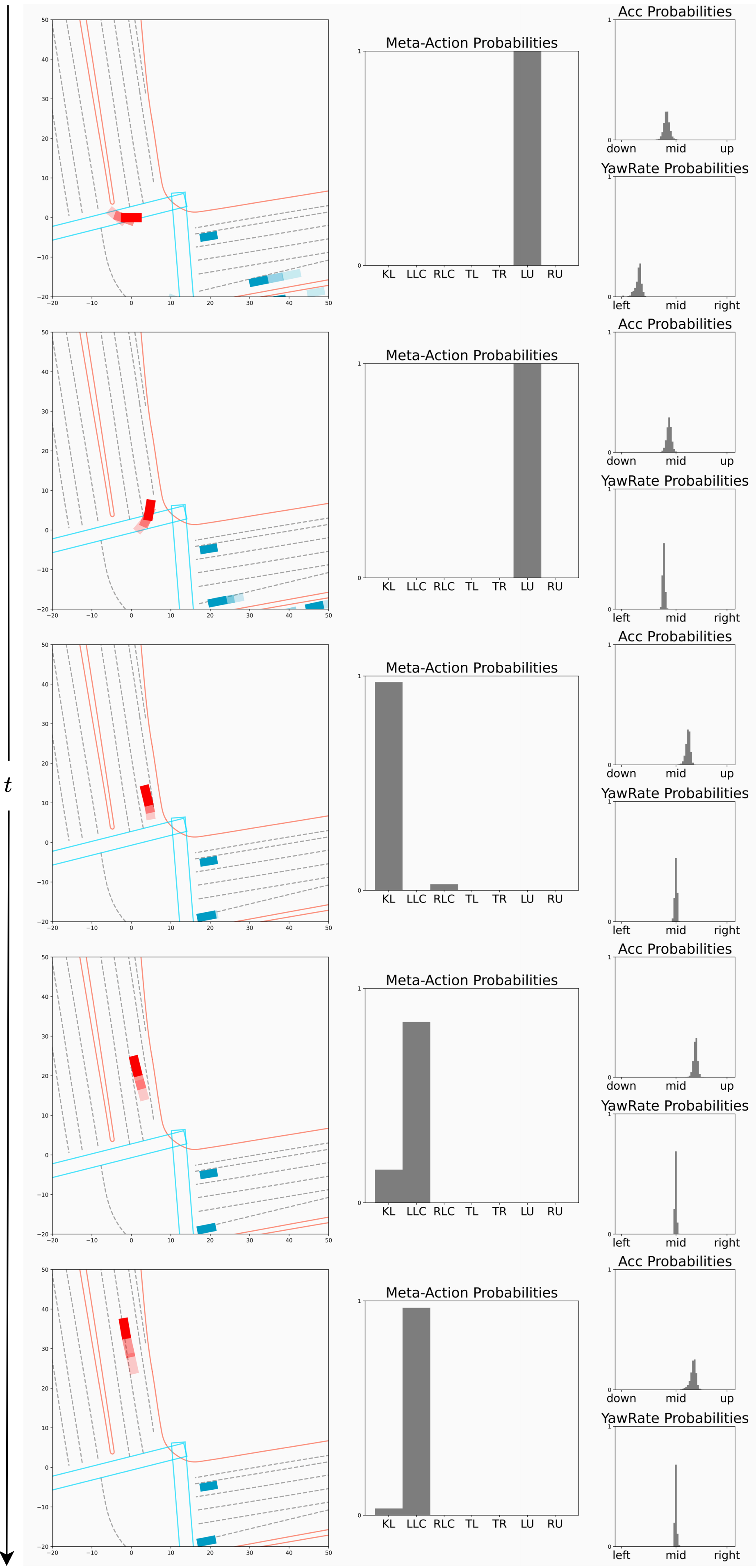}
\caption{Sampled rollout in a U-turn scenario: the ego vehicle first performs a Left U-turn (LU), briefly transitions to Keep Lane (KL), and subsequently initiates a Left Lane Change (LLC). This sequence reflects the model’s capacity for multi-stage behavioral planning through autoregressive meta-action inference.}
\label{fig:uturn_vis_0}
\end{figure}

\begin{figure}[!h]
\centering
\includegraphics[width=0.71\linewidth]{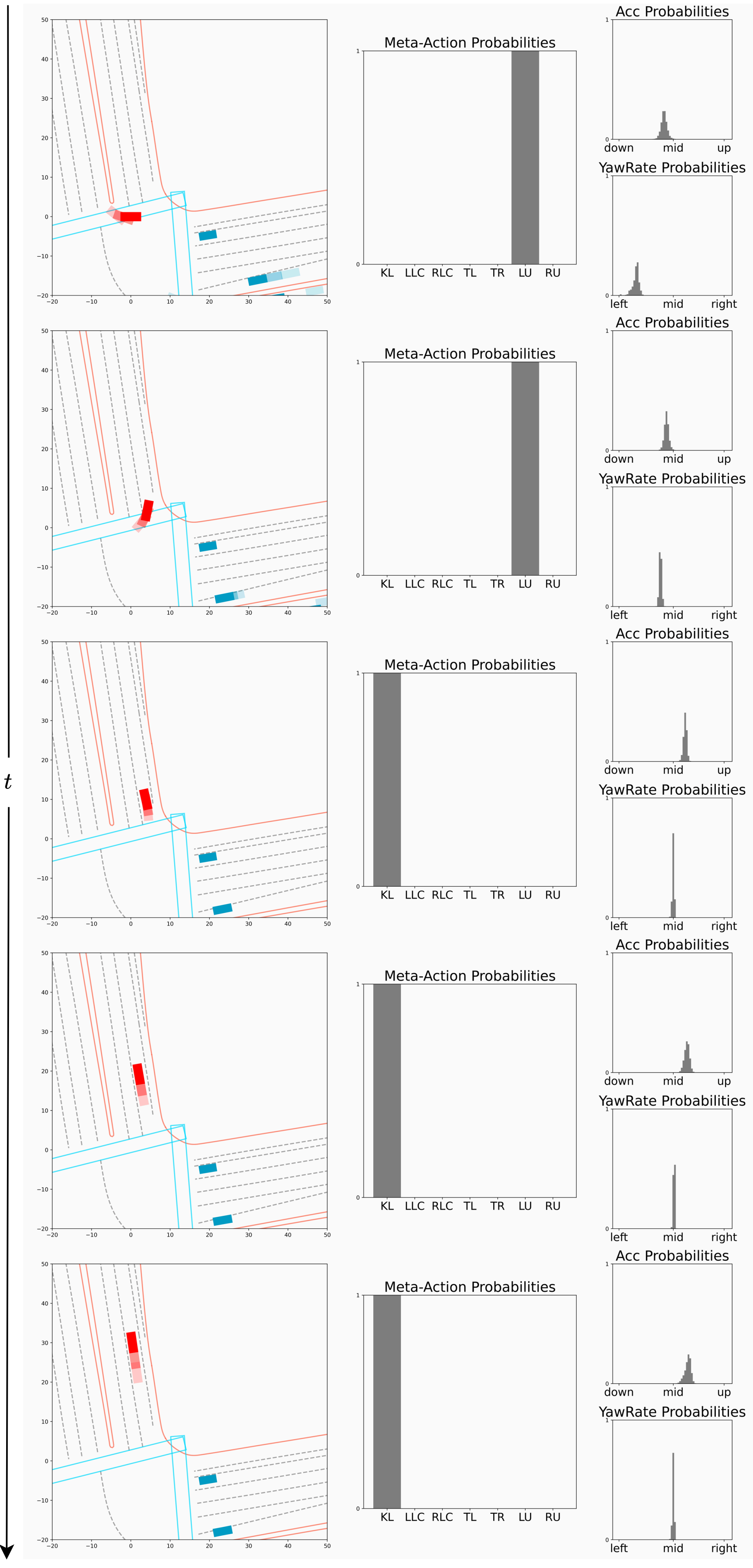}
\caption{Alternative sample from the same U-turn scenario: after completing the Left U-turn (LU), the ego vehicle continues in a Keep Lane (KL) state without further maneuvers. This case illustrates the model’s behavioral diversity enabled by stochastic meta-action prediction.}
\label{fig:uturn_vis_1}
\end{figure}

Next, we visualize a complex U-turn scenario from the validation set. In this case, we do not manually inject meta-actions; instead, meta-actions are sampled directly from the model’s predicted probability distribution. As shown in Figure~\ref{fig:uturn_vis_0}, the model first executes a left U-turn, briefly maintains a Keep Lane (KL) state, and subsequently transitions into a Left Lane Change (LLC). In contrast, Figure~\ref{fig:uturn_vis_1} presents an alternative sampled trajectory in which the ego vehicle continues straight after the U-turn without initiating a lane change. These variations demonstrate the model’s capacity to flexibly follow its own predicted meta-actions and emphasize the importance of autoregressive meta-action modeling in capturing diverse and context-adaptive behaviors in complex driving scenarios.

\newpage

\newpage

\section{Limitations and Broader Impact}
\label{sec:limitation}

\paragraph{Limitations.}
While our proposed framework introduces a unified and interpretable formulation for controllable trajectory generation, several limitations remain. First, the meta-action space is predefined and discrete, which may constrain the granularity and adaptability of behavior representation across diverse driving scenarios. In highly complex or ambiguous contexts, this fixed action vocabulary may be insufficient to capture nuanced driver intentions. Second, our meta-action labels are heuristically derived from trajectory data, which may introduce label noise or inconsistency—particularly near decision boundaries or during multi-agent interactions. Third, although our staged fine-tuning approach enables modular training, it may introduce suboptimal coordination between modules when new behaviors are added incrementally without re-training the foundation model.

\paragraph{Broader Impact.}
Our work contributes to the development of controllable and interpretable behavior models in autonomous driving, with the potential to improve transparency, testability, and decision accountability. By explicitly modeling high-level decision structures, our framework may support better human-in-the-loop validation, scenario-specific behavior control, and planning introspection. These benefits align with the broader goal of building trustworthy and verifiable intelligent systems. However, our approach also introduces the risk of oversimplifying decision processes by relying on coarse meta-actions, which could lead to undesired behavior in edge cases if not properly tested. Moreover, as with all predictive systems, careful consideration is needed to avoid over-reliance on learned patterns in safety-critical contexts.

\end{document}